\documentclass[letterpaper, 10 pt, conference]{ieeeconf} 

\IEEEoverridecommandlockouts               

\renewcommand{\baselinestretch}{0.99}
\usepackage{times}
\usepackage{epsfig}
\usepackage{graphicx}
\usepackage{amsmath}
\usepackage{amssymb}
\usepackage{bbding}
\usepackage{pifont}
\usepackage{wasysym}
\usepackage[table]{xcolor}
\usepackage{multirow}
\usepackage{pifont}
\newcommand{\cmark}{\ding{51}}%
\newcommand{\xmark}{\ding{55}}%

\usepackage{etoolbox}
\makeatletter
\patchcmd{\@makecaption}
 {\scshape}
 {}
 {}
 {}
\makeatother

\usepackage{array}
\newcolumntype{L}[1]{>{\raggedright\let\newline\\\arraybackslash\hspace{0pt}}m{#1}}
\newcolumntype{C}[1]{>{\centering\let\newline\\\arraybackslash\hspace{0pt}}m{#1}}
\newcolumntype{R}[1]{>{\raggedleft\let\newline\\\arraybackslash\hspace{0pt}}m{#1}}

\usepackage{cite}
 \usepackage{url}
 \usepackage{hyperref,xcolor}

\title{\LARGE \bf
Weakly Supervised Silhouette-based Semantic Scene Change Detection
}

\author{Ken Sakurada, Mikiya Shibuya, Weimin Wang 
\thanks{The authors are with Artificial Intelligence Research Center,
    National Institute of Advanced Industrial Science and Technology, Tokyo, Japan, 
    (e-mail: {\tt\scriptsize \{k.sakurada,mikiya-shibuya,weimin.wang\}@aist.go.jp})}%
}

\begin{document}

\maketitle
\thispagestyle{empty}
\pagestyle{empty}

\begin{abstract}

This paper presents a novel semantic scene change detection scheme with only weak supervision. A straightforward approach for this task is to train a semantic change detection network directly from a large-scale dataset in an end-to-end manner. However, a specific dataset for this task, which is usually labor-intensive and time-consuming, becomes indispensable. To avoid this problem, we propose to train this kind of network from existing datasets by dividing this task into change detection and semantic extraction. On the other hand, the difference in camera viewpoints, for example, images of the same scene captured from a vehicle-mounted camera at different time points, usually brings a challenge to the change detection task. To address this challenge, we propose a new siamese network structure with the introduction of correlation layer. In addition, we collect and annotate a publicly available dataset for semantic change detection to evaluate the proposed method. The experimental results verified both the robustness to viewpoint difference in change detection task and the effectiveness for semantic change detection of the proposed networks. Our code and dataset are available at \url{https://kensakurada.github.io/pscd}.

\end{abstract}

\section{INTRODUCTION}
Semantically understanding scene changes, such as semantic scene change detection, is one of the new problems that have attracted attention in the fields of computer vision, remote sensing, and natural language processing \cite{daudt2018fully,HRSCD2018,jhamtani2018learning,park2019viewpoint,suzuki2018semantic}.
Change detection methods have been comprehensively studied and applied to many kinds of tasks, such as detecting anomaly using surveillance and satellite cameras, inspecting infrastructure \cite{BMVC2015_127}, managing disaster \cite{Sakurada2013,sakurada2014massive}, and automating agriculture \cite{dong20174d}.
However, the existing methods of change detection specify a few detection targets, such as pedestrians and vehicles, for each application.
In cases where images contain various kinds of scene changes, more semantic information except for these targets is required for better discrimination in other advanced applications, such as updating city model for autonomous driving \cite{alcantarilla2016street}.

\begin{figure}[!t]
  \begin{center}
  \vspace{0mm}
  \hspace*{2mm}\includegraphics[width=75mm,bb=50 0 737 322]{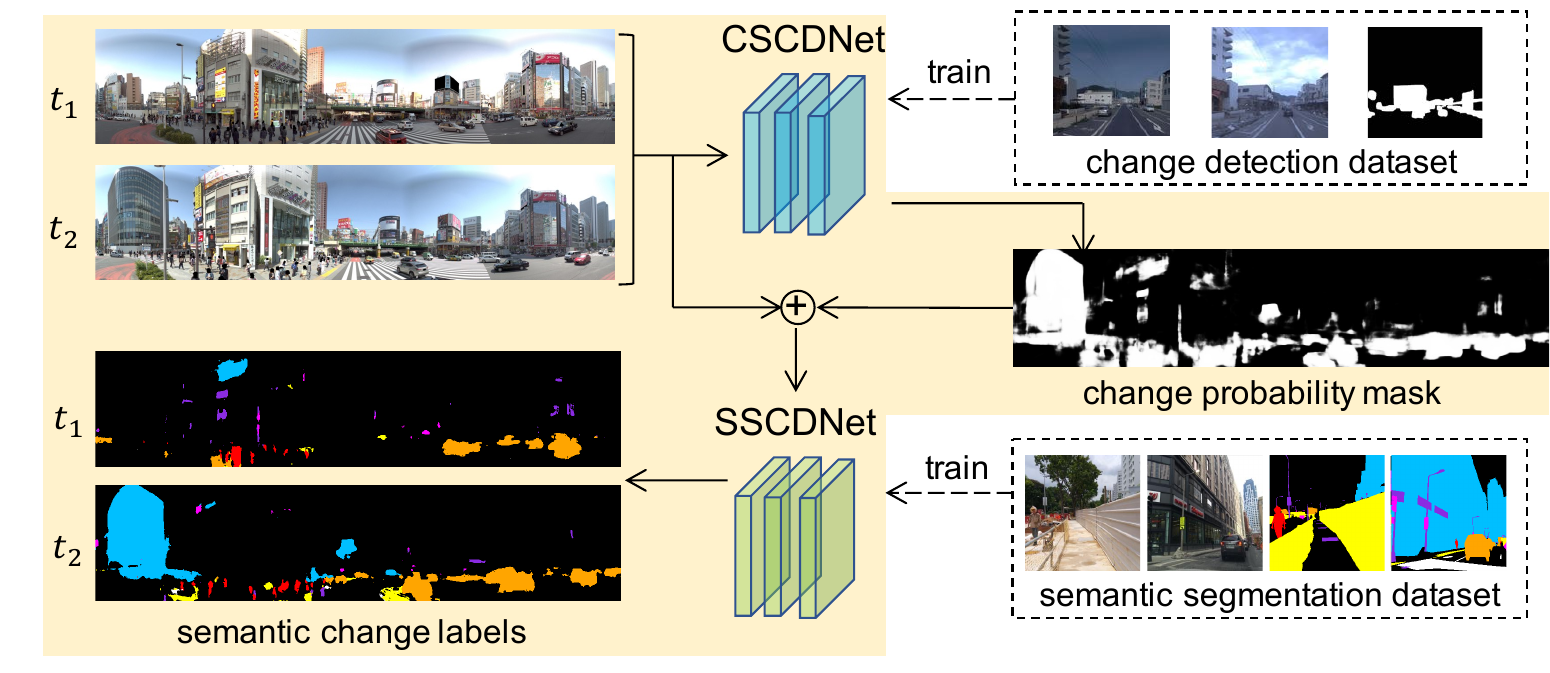}   
  \caption{Overview of the proposed method. First, the CSCDNet takes an image pair as input, which is trained using a change detection dataset, and outputs one change probability mask. Thereafter, the input image pair and the estimated change mask are fed into the SSCDNet, which is trained using a dataset synthesized from a semantic image segmentation dataset~\cite{Neuhold_2017_ICCV}. Finally, the SSCDNet estimates the pixel-wise semantic labels of each input image.}
  \label{fig:abst}
  \end{center}
  \vspace{-2mm}
\end{figure}

\begin{figure}[!t]
  \begin{center}
  $\begin{array}{p{27mm}p{27mm}p{27mm}}
  \hspace*{-1mm}\includegraphics[width=30mm,bb=0 0 200 202]{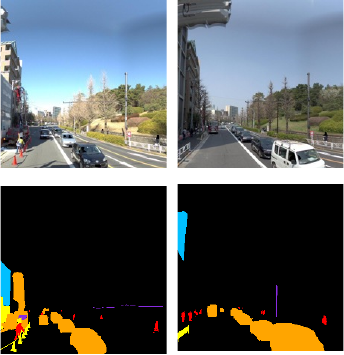}
  &\hspace*{-4mm}\includegraphics[width=30mm,bb=0 0 200 202]{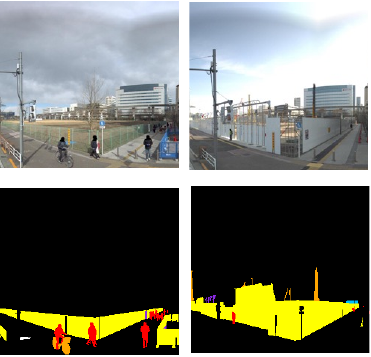}
  &\hspace*{-5mm}\includegraphics[width=30mm,bb=0 0 200 202]{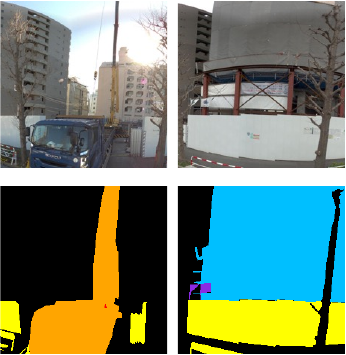} 
  \end{array}$
  $\begin{array}{p{27mm}p{27mm}p{27mm}}
    \hspace*{3mm}\raisebox{1mm}{\scriptsize (a) thing-to-thing}
    &\hspace*{1mm}\raisebox{1mm}{\scriptsize (b) stuff-to-stuff}
    &\hspace*{-1mm}\raisebox{1mm}{\scriptsize (c) thing-to-stuff}
  \end{array}$

  \caption{Examples of change types. (a) same class and different instance (cars to other ones), (b) same class {\it stuff} (barrier to other barrier), {\it thing-to-stuff} (vehicle to barrier). It should be noted that the definitions of {\it thing} and {\it stuff} (i.e. object instance) depends on the dataset and the application.}
  \vspace{-5mm}
  \label{fig:change_type}
  \end{center}
\end{figure}

Semantic scene change detection is a challenging task to detect and label scene changes on {\it each input image} (Fig. \ref{fig:abst}).
There are several types of scene changes in terms of {\it stuff and thing} classes \cite{kirillov2018panoptic} (Table \ref{table:change_type}). Figure \ref{fig:change_type} shows examples of each change type. 
One of the most straightforward methods is comparing the results of (I) pixel-wise semantic segmentation or (II) object detection between input images. Both of the methods (I) and (II) perform well for thing-to-thing changes of {\it different class}.
However, for example, in the case of thing-to-thing changes of {\it same class and different instance} such as Fig. \ref{fig:change_type} (a), the straightforward method fails to detect the instance changes.
The same thing applies to other change types (Fig. \ref{fig:change_type} (b), (c)).

Moreover, viewpoint changes of vehicular imagery are larger than those of images taken by surveillance and satellite cameras, which makes it complicated to detect changes between images with large variances of scene depth due to the problems of image correspondence, appearance change and occlusion.
Needless to say, although large-scale training datasets make it possible to estimate semantic changes with an end-to-end learning approach directly, 
it is labor-intensive to create large-scale semantic change detection datasets for each class definition of applications in terms of collecting and labeling images.

In order to overcome these difficulties, we propose a novel semantic change detection scheme with only weak supervision by dividing this task into change detection and semantic extraction (Fig.\ref{fig:abst}).
The proposed method is composed of the two convolutional neural networks (CNNs), a correlated siamese change detection network (CSCDNet), and a silhouette-based semantic change detection network (SSCDNet).
First, the CSCDNet takes an image pair as input and outputs one change probability mask. Thereafter, the input image pair and the estimated change mask are fed into the SSCDNet. Finally, it estimates the pixel-wise semantic labels of each input image.

The SSCDNet can be trained with the dataset synthesized from commonly available semantic image segmentation datasets, such as the Mapillary Vistas dataset \cite{Neuhold_2017_ICCV}, to avoid creating a new dataset for semantic change detection.
The estimation accuracy of the SSCDNet depends on that of change detection.
However, in the case of images captured from a vehicle-mounted camera at different time points, existing change detection methods suffer from estimation errors due to differences in camera viewpoints.
Hence, we propose a new siamese network architecture with the introduction of correlation layers, named as the CSCDNet, which is trained using a change detection dataset.
The CSCDNet can deal with differences in camera viewpoints and achieves state of the art performance on the panoramic change detection (PCD) dataset \cite{sakurada2015change}.
Additionally, we incorporate the data augmentation for the input change mask in the training step to improve the robustness of the SSCDNet to change detection errors.
For evaluating the proposed methods, we have created the panoramic semantic change detection (PSCD) dataset in the hopes of accelerating researches in the field of dynamic scene modeling.

Our main contributions are as follows:
\renewcommand{\labelitemi}{$\bullet$}
\begin{itemize}
  \vspace{-.5mm}
  \item We propose a novel semantic change detection network that can be trained with only weak supervision from existing datasets.
  \vspace{-.5mm}
  \item Our siamese change detection network, which uses correlation layers that can deal with differences in camera viewpoints, achieves state of the art performance on the PCD dataset.
  \vspace{-.5mm}
  \item We create the first publicly available street-level image dataset for semantic scene change detection.
\end{itemize}

This paper is organized as follows. In Sec.\ref{sec:related_work}, we summarize the related work. 
Section \ref{sec:proposed_method} explains the details of the proposed network and the training method.
Section \ref{sec:experiment} shows the experimental results. 
Section \ref{sec:conclusion} presents our conclusions.

\begin{table}[!t]
  \caption{Change types and the applicability of each semantic change detection method. The straightforward methods (I) and (II) are comparing estimated labels of (I) pixel-wise semantic segmentations and (II) object detections, respectively.}
  \vspace{0mm}
  \label{table:change_type}
  \centering
  {\footnotesize
  \begin{tabular}{c|c|c|c|c}
    \hline
     \multicolumn{2}{c|}{\multirow{2}{*}{change type} }& \multicolumn{3}{c}{method} \\ 
    \cline{3-5}
     \multicolumn{2}{c|}{} & (I) & (II) & ours \\ 
    \hline
    \hline
    \multirow{2}{*}{thing-to-thing} & different class & \cmark & \cmark & \cmark  \\ \cline{2-5}
     & same class \& different instance & \xmark & \xmark & \cmark  \\
    \hline
    \multirow{2}{*}{stuff-to-stuff} & different class & \cmark & \xmark & \cmark  \\ \cline{2-5}
     & same class  & \xmark & \xmark & \cmark  \\
    \hline
    thing-to-stuff & \multirow{2}{*}{different class} & \multirow{2}{*}{\cmark} & \multirow{2}{*}{\xmark} & \multirow{2}{*}{\cmark}  \\ 
    stuff-to-thing & & & & \\
    \hline
  \end{tabular}
  }
\end{table}

\section{Related Work}
\label{sec:related_work}
Many methods for temporal scene modeling have been proposed.
However, most of them focused on detecting changes or estimating the length of time that each part of a scene exists for.
Semantic recognition is required for advanced applications based on dynamic modelings, such as autonomous driving and augmented reality.
This section explains the reason for the proposal of the semantic change detection method using commonly available semantic image segmentation datasets.

 \subsection*{Change Detection}
 \label{subsec:relatede_work_change_detection}
Change detection methods are classified into several categories depending on types of target scene changes and available information.
Change detection in 2D (image) domain is the most standard approach, especially for surveillance and satellite cameras \cite{crispell2012variable,Huertas1998,Pollard2007,Radke2005}, which are accurately aligned. 
A typical approach models the appearance of the scene from a set of images captured at different times, against which a newly captured query image is compared to detect changes \cite{wang2018m4cd}.

Some studies formulate the problem in a 3D domain.
Schindler et al. proposed the probabilistic temporal inferences model based on the visibility of each 3D point reconstructed from images taken from multiple viewpoints at different times \cite{Schindler2010}.
The work by Matzen et al. \cite{MatzenECCV14} is classified into the same category.
In terms of application, the works by Taneja et al. \cite{Taneja2011,Taneja2013}, and Sakurada et al. \cite{Sakurada2013} might be the closest to our research.

In recent years, significant efforts have been made to change detection using machine learning, especially for deep neural networks (DNNs) \cite{BMVC2015_127,alcantarilla2016street,sakurada2015change,fujita2017damage,khan2017learning}.
There are mainly two types of formulations, ``patch similarity estimation" and ``pixel-wise segmentation".
Patch similarity estimation has been studied for not only change detection but also feature, stereo, and image matchings \cite{chen2015deep,lin2015learning,simo2015discriminative,zagoruyko2015learning,zbontar2016stereo}.
Pixel-wise change detection has been further studied in the context of anomaly detection, background subtraction, and moving object detection \cite{khan2017learning,camplani2017benchmarking,toyama1999wallflower}.

\subsection*{Semantic Change Detection}
\label{subsec:relatede_work_semantic_change_detection}
There are few studies on semantic change detection because most of change detection studies that specify their target domain, such as moving object, forest, and do not explicitly recognize semantic classes of change.
The work by Suzuki et al. \cite{suzuki2018semantic} proposes a method to classify a change mask using multi-scale feature maps extracted using a CNN. It does not consider the problem of detecting changes and estimating correspondences between input images and the change mask (e.g., in Fig.\ref{fig:change_type}, there are different change regions between two input images).
Daudt et al. \cite{daudt2018fully,HRSCD2018} detected land surface changes between satellite images.
In the case of land surface change detection of satellite images, unlike scene change detection, it is unnecessary to estimate correspondences between input images and the change mask because the change regions between the input images are common. 
However, for street-level scene change detection, the estimation is necessary because scene objects can appear, disappear, and move.  

\section{Weakly Supervised Silhouette-based Semantic Scene Change Detection}
\label{sec:proposed_method}
There are many types of label definitions for semantic image segmentation depending on the applications; for example, ground-level images of indoor and outdoor scenes \cite{Neuhold_2017_ICCV,armeni2017joint}, aerial and satellite images \cite{ISPRS_semantic_contest,MnihThesis}.
Additionally, the definition of {\it change} (e.g., whether changes of moving objects, display of digital screens, the light of a lamp, transparent barriers, growth of plants, a pool of water, and seasonal changes of vegetation are ignored or not) depends on the application.
Thus, there is a large number of combinations of change and semantic definitions.
Clearly, it is time-consuming to create semantic change detection datasets for each application.
Furthermore, as mentioned above (Sec.\ref{sec:related_work}), it is necessary to estimate correspondences between input images and the change mask because the existing change detection datasets do not explicitly contain that information.

To solve these problems, the proposed method includes two CNNs, namely, the CSCDNet and the SSCDNet. 
This separated architecture enables the method to train the semantic change detection system with change detection datasets and commonly available semantic image segmentation datasets.
The rest of this section explains the details regarding the weakly supervised method.

\begin{table}[!t]
  \caption{Details of the datasets used in the experiments. \scriptsize{*(The CSCDNet is trained with only image pairs of a scene and their change masks of the PSCD dataset.)}}
  \vspace{0mm}
  \label{table:datasets}
  \centering
  {\footnotesize
  \begin{tabular}{c|c|c|c|c}
    \hline
     \multirow{2}{*}{Dataset} &   \multicolumn{2}{c|}{PCD \cite{sakurada2015change}} & \multirow{2}{*}{Vistas \cite{Neuhold_2017_ICCV}}  & PSCD \\
     \cline{2-3}
     & \hspace{-2mm} TSUNAMI \hspace{-2mm} & \hspace{-1mm} GSV \hspace{-1mm} & & (This work) \\
    \hline
    \hline
    Number of images  & 100 & 100 & 20,000 & 770  \\
    \hline
    Original size &  \multicolumn{2}{c|}{1024 $\times$ 224} & various & 4096 $\times$ 1152  \\
    \hline
    Crop size &  \multicolumn{2}{c|}{224 $\times$ 224} & - & 224 $\times$ 224  \\
    \hline
     Size in training &  \multicolumn{2}{c|}{256 $\times$ 256} & 256 $\times$ 256 & 256 $\times$ 256  \\
    \hline
    Paired & \multicolumn{2}{c|}{\checkmark} & - & \checkmark  \\
    \hline
    Change mask & \multicolumn{2}{c|}{\checkmark} & - & \checkmark  \\
    \hline
    Semantic label & \multicolumn{2}{c|}{-} & \checkmark & \checkmark  \\
    \hline     
    Alignment & medium & \hspace{-2mm} coarse \hspace{-2mm} & - & coarse  \\
    \hline
    Training target & \multicolumn{2}{c|}{CSCDNet} & SSCDNet & CSSCDNet   \\
     & \multicolumn{2}{c|}{} & & *(CSCDNet) \\
    \hline
  \end{tabular}
  }
\end{table}

\subsection{Overview}
Figure \ref{fig:abst} shows an overview of the proposed semantic change detection method.
First, the CSCDNet takes an image pair as input, which is trained using a change detection dataset and outputs the change probability of each pixel as one change mask image.
Subsequently, the input image pair and the estimated change mask are fed into the SSCDNet, which is trained using a dataset synthesized from a semantic image segmentation dataset.
Finally, the SSCDNet estimates the pixel-wise semantic labels of each input image.
It should be noted here that the SSCDNet can estimate semantic change labels and correspondences between input images and the change mask simultaneously. 

We conjecture that these semantic label estimations and splitting the change mask into the input images can be trained using a commonly available semantic image segmentation dataset, such as the Mapillary Vistas dataset \cite{Neuhold_2017_ICCV}, and that semantic information can improve the accuracy of the change mask estimation. 
Table \ref{table:datasets} shows the details of the datasets used in this paper.
The experimental results show the effectiveness of this strategy (Sec.\ref{sec:experiment}). 
The details of the training dataset synthesis and the network architectures are explained in the following subsections.

\subsection{Dataset Synthesis from Semantic Segmentation Dataset}
\label{subsec:training_datatset}
Here, we consider the problem of estimating pixel-wise semantic change labels of each input image from an image pair and the change mask.
There are several possible methods for generating training datasets to solve this problem.
A simulator using a photorealistic rendering, such as Virtual KITTI \cite{Gaidon:Virtual:CVPR2016}, SYNTHIA \cite{ros2016synthia} and SceneNet RGB-D \cite{McCormac:etal:ICCV2017} datasets, is one solution.
Although photorealistic images might be effective for pre-training, fine-tuning is necessary to address the domain gaps between synthetic and real images.
To bridge the gap, Shrivastava et al. proposed the method to learn a model to improve the realism of a simulator's output using unlabeled real data \cite{shrivastava2017learning}. However, it is difficult to directly apply this method to natural scene images, which are more complicated than their target domains.
Alternatively, synthesis using real images can be applied.
Dwibedi et al. proposed the synthetic method to generate large annotated instance datasets in a cut and paste manner \cite{Dwibedi_2017_ICCV}.
Their study might be the closest to our method.

\begin{figure}[!t]
  \begin{center}
  \vspace{0mm}
  \hspace*{-1mm}\includegraphics[width=85mm,bb=0 0 834 235]{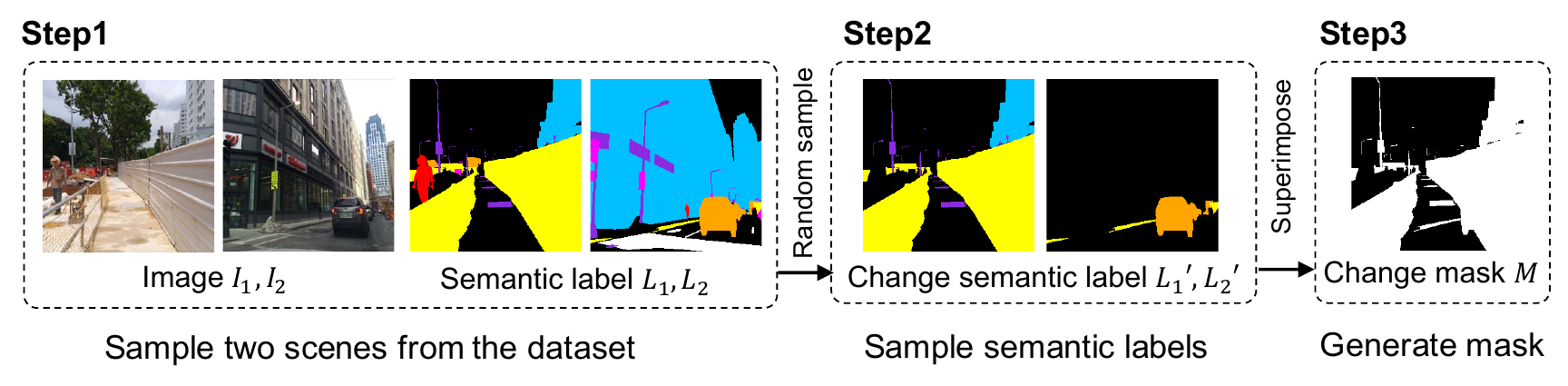}   
  \caption{Synthesis of training dataset for the SSCDNet from semantic image segmentation dataset~\cite{Neuhold_2017_ICCV}.}
  \label{fig:synthesis}
  \vspace{-2mm}
  \end{center}   
\end{figure}

Figure \ref{fig:synthesis} shows an overview of the proposed training dataset synthesis for the SSCDNet from a semantic image segmentation dataset.
First, two RGB images $I_1,I_2$, and their semantic label images $L_1,L_2$ are randomly sampled from the semantic image segmentation dataset.
Thereafter, the change semantic label images ${L_1}',{L_2}'$ are generated by sampling $n$ semantic labels randomly and removing the others from each semantic label image $(1 \leq n \leq \min (n_\mathrm{max},N_i-1))$.
$N_i$ represents the number of the classes that the semantic label image $L_i$ contains.
The maximum number of class samplings $n_\mathrm{max}$ should be decided depending on the number of classes of the semantic segmentation dataset.
Finally, the change mask is generated by superimposing the randomly sampled semantic labels as binary silhouettes $M$.

\begin{figure}[!t]
  \begin{center}
  \vspace{0mm}
  \hspace*{-2mm}\includegraphics[width=80mm,bb=0 0 500 350]{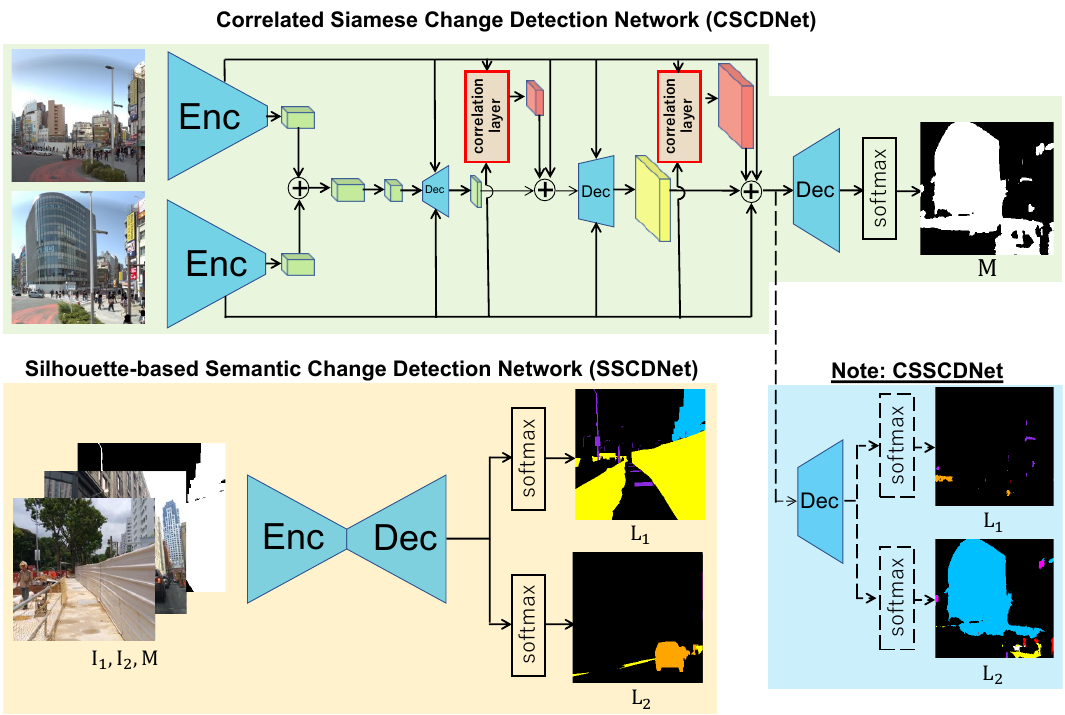}   
  \caption{Network architectures of the CSCDNet, the SSCDNet, and the CSSCDNet. The architecture of the CSSCDNet is based on the CSCDNet and its output layer is replaced with that of the SSCDNet. (The images $I_1$, $I_2$, $M$, $L_1$ and $L_2$ in the SSCDNet are parts of the Vistas dataset \cite{Neuhold_2017_ICCV}.)} 
  \label{fig:CNN_architecture}
  \vspace{-2mm}
  \end{center}   
\end{figure}

\subsection{Network Architecture}

\hspace{-5mm} \textbf{\textup{Correlated Siamese Change Detection Network (CSCDNet) } }: We propose the CSCDNet to overcome the limitation of the camera viewpoints of the previous methods.
Figure \ref{fig:CNN_architecture} shows an overview of the network architecture of the proposed method.
As mentioned in Sec.\ref{sec:related_work}, Sakurada et al. \cite{sakurada2015change} found that the comparison between feature maps extracted from input images using a CNN trained with large-scale image recognition datasets \cite{sakurada2017dense} is effective for scene change detection task.
To incorporate this advantage, we chose the siamese network architecture based on the ResNet-18 \cite{he2016deep} which was pretrained on the ImageNet \cite{imagenet_cvpr09} dataset as the encoder of the CSCDNet.
Each feature map extracted from two input images in the encoder is concatenated with each decoder's output and fed into the next layer of the decoder whose architecture is based on the network by \cite{Hamaguchi_2018_CVPR_Workshops}.

Furthermore, for the situation of an image pair with a large viewpoint difference, this difference has to be considered in the design of the network structure to improve the detection accuracy.
Exploiting the dense optical flow estimated by the other methods \cite{sakurada2017dense} is not efficient in terms of optimization.
Therefore, we inserted correlation layers \cite{Dosovitskiy_2015_ICCV}, which are utilized for the estimation of optical flow and stereo matching, into the siamese network.

The CSCDNet takes images $I_1$ and $I_2$ captured at times $t_1$ and $t_2$ as an input.
Each pixel value is normalized in $[-1,1]$.
The change mask, as the ground-truth, $M_g$, is provided to the output of the network as training data.
After the final convolution layer, the feature maps are evaluated by the following pixel-wise binary cross-entropy loss:
\begin{equation}
  \mathcal{L}_{c}=-\sum_{\mathbf{x}} t(\mathbf{x}) \ln (p_c(\mathbf{x})) + (1-t(\mathbf{x})) \ln (1-p_c(\mathbf{x})),
\end{equation}
where $\mathbf{x}$, $t(\mathbf{x})$ and $p_c(\mathbf{x})$ represent the pixel coordinates of the output change mask, the ground-truth, and predictions computed using each output feature maps by a pixel-wise softmax, respectively.

\vspace{2mm}
\hspace{-5mm} \textbf{\textup{Silhouette-based Semantic Change Detection Network (SSCDNet)}} : The architecture of the SSCDNet is based on the combination of U-Net based on ResNet-18 \cite{he2016deep,Hamaguchi_2018_CVPR_Workshops}.
Their main differences are the input and output parts.
The SSCDNet takes images $I_1$, $I_2$ and $M$, which are concatenated in the channel dimension as a seven-channel image, for the input.
Moreover, after the final convolution layer, the output feature maps are split in half (the bottom of Fig.\ref{fig:CNN_architecture}), and each of the feature maps is evaluated by the following pixel-wise cross-entropy loss:
\begin{equation}
  \mathcal{L}_{s}=-\sum_{\mathbf{x}}\sum_{k} t_1(\mathbf{x},k) \ln (p_1(\mathbf{x},k)) + t_2(\mathbf{x},k) \ln (p_2(\mathbf{x},k)),
\end{equation}
where $k$ is an index of classes ($1\leq k \leq K$, $K$: the number of classes), $t(\mathbf{x},k)$ represents the ground-truth with 1-of-$K$ coding scheme, $p(\mathbf{x},k)$ represents predictions computed from each output feature maps by a pixel-wise softmax.

\vspace{2mm}
\hspace{-5mm} \textbf{\textup{Correlated Siamese Semantic Change Detection Network (CSSCDNet)}} : For a comparative study, we proposed the CSSCDNet as a naive method in the case that the semantic change detection dataset is available.
The architecture is based on the CSCDNet.
After the final convolution layer, the output feature maps are split in half, and each of the feature maps is evaluated by the pixel-wise cross-entropy loss in the same manner as the SSCDNet (in the dash line box of Fig.\ref{fig:CNN_architecture}).

\begin{table}[t]
  \caption{$\mathrm{F_1 score}$ and mIoU of change detection for TSUNAMI and GSV datasets. Siamese-CDResNet represents the CSCDNet without correlation layers. The CSCDNet consistently outperforms the other methods. }
  \label{table:score_cdnet}
  \centering
  {\footnotesize
  \vspace{1mm}
  \begin{tabular}{c|ccc}
    \hline
        & &$\mathrm{F_1 score}$ (mIoU) & \\
    \hline
        & TSUNAMI & GSV & Average \\
    \hline \hline
    DenseSIFT \cite{sakurada2015change} & 0.649 (--) & 0.528 (--)& 0.589 (--) \\
    \hline
    CNN-feat \cite{sakurada2015change} & 0.723 (--) & 0.639 (--) & 0.681 (--) \\
    \hline
    DeconvNet \cite{alcantarilla2016street} & 0.774 (--) & 0.614 (--) & 0.694 (--) \\
    \hline
    WS-Net \cite{khan2017learning} & -- (--) & -- (--) & -- (0.477) \\
    \hline
    FS-Net \cite{khan2017learning} & -- (--) & -- (--) & -- (0.588) \\
    \hline
    CDNet \cite{sakurada2017dense} & 0.848 (0.811) & 0.695 (0.672) & 0.772 (0.741) \\
    \hline
     \multirow{2}{*}{\shortstack[c]{CosimNet-\\ 3layer-l2 \cite{guo2018learning}} } & \multirow{2}{*}{0.806 (--)} & \multirow{2}{*}{0.692 (--)} & \multirow{2}{*}{0.749 (--)} \\ \\
    \hline
     \multirow{2}{*}{\shortstack[c]{Siamese-\\ CDResNet (Ours) }} & \multirow{2}{*}{0.850 (0.815)} & \multirow{2}{*}{0.718 (0.691)} & \multirow{2}{*}{0.784 (0.753)} \\ \\
    \hline
    CSCDNet (Ours) & \textbf{0.859} (\textbf{0.824}) & \textbf{0.738} (\textbf{0.706}) & \textbf{0.799} (\textbf{0.765}) \\ 
    \hline

  \end{tabular}
  }
\vspace{-2mm}
\end{table}

\section{Experiments}
\label{sec:experiment}
To evaluate the effectiveness of our approach, we performed three experiments.
The first experiment is the accuracy evaluation of the change detection with the CSCDNet on the PCD dataset \cite{sakurada2015change}.
The proposed siamese change detection networks with and without correlation layers and other existing methods are compared.
The second experiment is an accuracy evaluation of the semantic change detection with the SSCDNet using datasets synthesized from the Mapillary Vistas dataset \cite{Neuhold_2017_ICCV}.
The data augmentation of the change mask is also evaluated, which improves the robustness of the SSCDNet against change detection errors of the CSCDNet.
In the final experiment, we applied our semantic change detection method to the PSCD dataset, which is different from the training dataset of the SSCDNet, and show the effectiveness of our approach.

\subsection{Panoramic Semantic Change Detection (PSCD) dataset}
For the quantitative evaluation of the proposed approach, we have created a new dataset named the PSCD dataset, which opens up new vistas for semantic change detection.
The PSCD dataset comprises 770 panoramic image pairs. 
Each pair consists of images $I_1, I_2$ taken at two different time points $t_1$, and $t_2$.

The PSCD dataset contains the change binary masks $C_{1}, C_{2}$, the intersection change mask $C_{1 \& 2}$, the semantic labels $S_1,S_2$, the instance labels $D_1, D_2$, the attributes $A_{1}, A_{2}$ (3D object, 2D texture), the privacy masks $P_{1}, P_{2}$, the intersection privacy mask $P_{1 \& 2}$. 
We defined the 67 semantic classes based on those of the Mapillary Vistas dataset \cite{Neuhold_2017_ICCV}, and integrated the original classes into the $N=8$ classes based on the map updating applications as shown in Fig.\ref{fig:result_pscd} \footnote{See the details on the project page \url{https://kensakurada.github.io/pscd}.}.

\subsection{Experimental Settings}

\subsubsection{Training dataset generation}
We generated training datasets for the CSCDNet, the SSCDNet, and the CSSCDNet from the PCD, the Mapillary Vistas, and the PSCD datasets, respectively.
Table \ref{table:datasets} shows the details of the dataset.
The PCD dataset is composed of panoramic image pairs $I_1$, $I_2$ taken at two different time points $t_1$, and $t_2$, and the change mask $M_g$.
From the image set $[I_1,I_2,M_g]$, patch images are cropped by sliding and resized.
Furthermore, data augmentation is performed by rotating the patches.
Thus, 12,000 sets of image patches were generated.
The PSCD dataset is resized and cropped, and data augmentation is performed in the same way as the PCD dataset.

We also generated training datasets for the SSCDNet from the Mapillary Vistas dataset \cite{Neuhold_2017_ICCV}.
The Mapillary Vistas dataset for research use contains 20,000 scene images and the pixel-wise semantic labels with 66 semantic classes (including an unlabeled class).
We integrate them into the following 8 classes: {\it no change, vehicles, barrier, structure, lane marking, object (traffic), object (others), human}. 
We selected the value of $n_\mathrm{max}$ as $N-1=7$ based on the ablation study. 
Figure \ref{fig:synthesis} shows an example of the dataset synthesized by the proposed method.

\subsubsection{Data augmentation for robustness to change detection error}
\label{subsec:mask_da}
If the change masks that are synthesized from semantic segmentation datasets are directly used in the training of the SSCDNet, the trained SSCDNet can be vulnerable to errors in change detection.
To improve the robustness of the SSCDNet to change detection error, we perform the data augmentation for change mask in training.
Specifically, the change mask is randomly applied to one of the four morphological transformations (erosion, dilation, opening, closing) with a random kernel size $k$ ($1 \leq k \leq 20$).
We expect that the semantic label information can reduce the error of semantic change detection due to the error of change detection by simulating the change mask.

\subsubsection{Training details}
The CSCDNet, the SSCDNet and the CSSCDNet are trained using four Nvidia V100 GPUs (CSCDNet in the change detection experiment with the PCD dataset is trained using eight Nvidia Tesla P100).
The networks are implemented using the PyTorch framework. 
We used the batch size of 32. 
The numbers of iteration for the CSCDNet, the SSCDNet, and the CSSCDNet are $3\times10^4$, $1\times10^5$, and $1\times10^5$, respectively. 
The Adam algorithm, with a learning rate of $2 \times 10^{-4}$, is used. 
The evaluations of the estimation accuracies of the CSCDNet using the PCD dataset and the CSSCDNet using the PSCD dataset are performed using the five-fold cross-validation.

\subsection{Evaluation}

\subsubsection{Change detection for the PCD dataset}
Table \ref{table:score_cdnet} shows $F_{1}$ scores and mean intersection-over-union (mIoU) of each method for TSUNAMI and GSV datasets.
The CSCDNet outperforms the other methods in terms of both $F_{1}$ scores and mIoU.
Furthermore, the improvements in the scores for GSV are more significant than those of TSUNAMI.
The main reason is that GSV contains more precise changes and the camera viewpoint differences are relatively larger than TSUNAMI because of the differences in their scene depths.
The CSCDNet can accurately detect the precise scene changes dealing with the differences in camera viewpoints.

\begin{figure}[!t]
  \begin{center}
  \vspace{0mm}
  $\begin{array}{p{9mm}p{9mm}p{9mm}p{9mm}p{9mm}p{9mm}p{9mm}}
  \hspace*{-1mm}\includegraphics[width=11mm,bb=0 0 256 256]{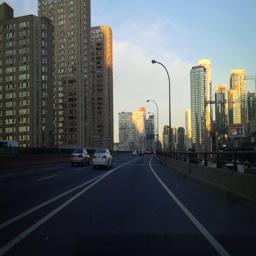}
  &\hspace*{-2mm}\includegraphics[width=11mm,bb=0 0 256 256]{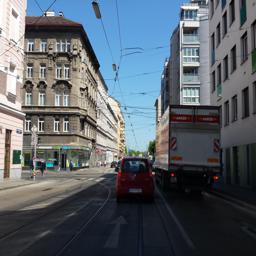}
  &\hspace*{-3mm}\includegraphics[width=11mm,bb=0 0 256 256]{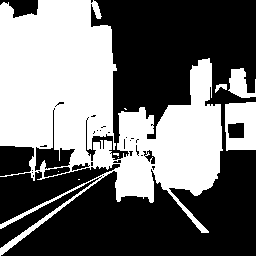} 
  &\hspace*{-3mm}\includegraphics[width=11mm,bb=0 0 256 256]{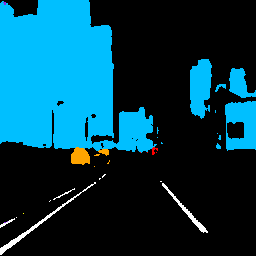} 
  &\hspace*{-4mm}\includegraphics[width=11mm,bb=0 0 256 256]{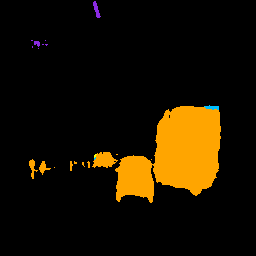} 
  &\hspace*{-4mm}\includegraphics[width=11mm,bb=0 0 256 256]{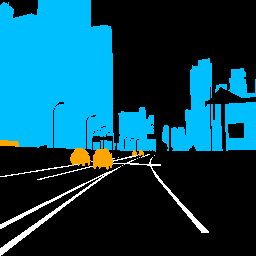} 
  &\hspace*{-5mm}\includegraphics[width=11mm,bb=0 0 256 256]{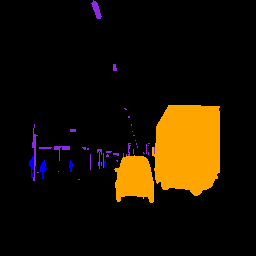} 
  \end{array}$
  $\begin{array}{p{27mm}p{18mm}p{18mm}}
    \hspace*{-4mm}\raisebox{1mm}{\scriptsize Input images and change mask}
    &\hspace*{5mm}\raisebox{1mm}{\scriptsize Estimation}
    &\hspace*{6mm}\raisebox{1mm}{\scriptsize Ground-truth}
  \end{array}$
  \vspace{-2mm}
  \caption{Example of results estimated by the SSCDNet. (The input and the ground-truth images are parts of the Vistas dataset \cite{Neuhold_2017_ICCV}.)}
  \vspace{-2mm}
  \label{fig:sscdnet_test}
  \end{center}
\end{figure}

\begin{table}[t]
  \caption{mIoU of SSCDNet for synthetic data from Mapillary Vistas dataset. }
  \label{table:iou_sscdnet}
  \centering
  {\small
  \begin{tabular}{c|C{1cm}C{1cm}|C{1cm}C{1cm}}
    \hline
    DA for test &  \multicolumn{2}{|c|}{-} &\multicolumn{2}{|c}{\checkmark} \\
    \hline
    DA for training & - & \checkmark & - & \checkmark  \\
    \hline
    \hline
    \textbf{mIoU} & \textbf{0.580} & 0.509 & 0.364 & \textbf{0.432} \\
    \hline
  \end{tabular}
  }
  \vspace{2mm}
\end{table}

\begin{figure}[!t]
  \begin{center}
  $\begin{array}{p{10mm}p{10mm}p{10mm}p{10mm}p{10mm}p{10mm}}
    \hspace*{-1.5mm}\includegraphics[width=11mm,bb=0 0 256 256]{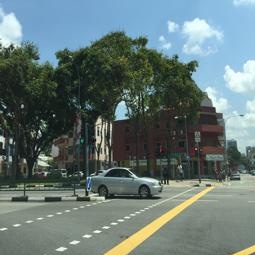}
    &\hspace*{-2.5mm}\includegraphics[width=11mm,bb=0 0 256 256]{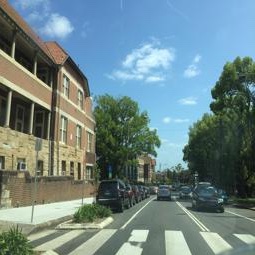}
    &\hspace*{-0.5mm}\includegraphics[width=11mm,bb=0 0 256 256]{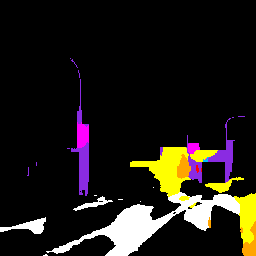} 
    &\hspace*{-1.5mm}\includegraphics[width=11mm,bb=0 0 256 256]{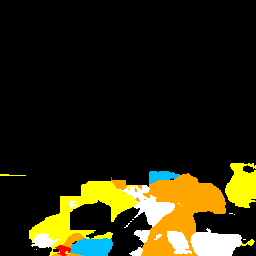} 
    &\hspace*{0.5mm}\includegraphics[width=11mm,bb=0 0 256 256]{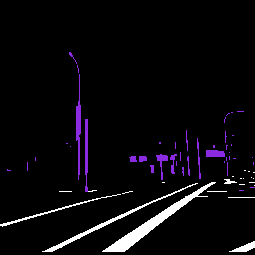} 
    &\hspace*{-0.5mm}\includegraphics[width=11mm,bb=0 0 256 256]{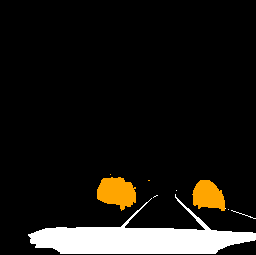} 
  \end{array}$

  $\begin{array}{p{10mm}p{10mm}p{24mm}p{10mm}p{10mm}}
    \hspace*{4mm}\raisebox{1mm}{\scriptsize $I_1$}
    &\hspace*{2mm}\raisebox{1mm}{\scriptsize $I_2$}
    &\hspace*{2.5mm}\raisebox{1mm}{\scriptsize Trained without DA}
    &\hspace*{5mm}\raisebox{1mm}{\scriptsize $L_1$}
    &\hspace*{4mm}\raisebox{1mm}{\scriptsize $L_2$}
  \end{array}$

  $\begin{array}{p{10mm}p{10mm}p{10mm}p{10mm}p{10mm}p{10mm}}
    \hspace*{5mm}\includegraphics[width=11mm,bb=0 0 256 256]{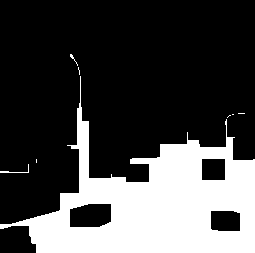}
    &
    &\hspace*{-0.5mm}\includegraphics[width=11mm,bb=0 0 256 256]{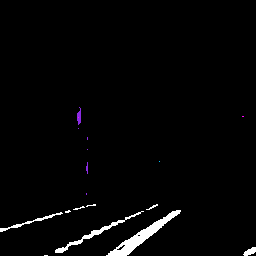} 
    &\hspace*{-1.5mm}\includegraphics[width=11mm,bb=0 0 256 256]{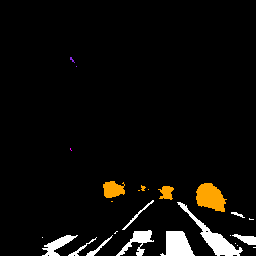} 
    & \hspace*{7mm}\includegraphics[width=11mm,bb=0 0 256 256]{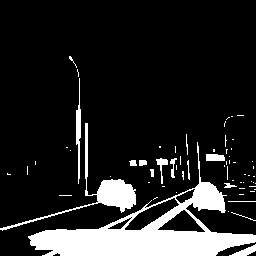} 
    &
  \end{array}$

  $\begin{array}{p{22mm}p{22mm}p{22mm}}
    \hspace*{-5.5mm}\raisebox{1mm}{\scriptsize Change mask with noise $M'$}
    &\hspace*{3mm}\raisebox{1mm}{\scriptsize Trained with DA}
    &\hspace*{6mm}\raisebox{1mm}{\scriptsize Change mask $M$}
  \end{array}$

  \vspace{1mm}

  $\begin{array}{p{22mm}p{22mm}p{22mm}}
    \hspace*{1mm}\raisebox{1mm}{\small Input images}
    &\hspace*{-4mm}\raisebox{1mm}{\small Predictions by SSCDNet}
    &\hspace*{6mm}\raisebox{1mm}{\small Ground-truth}
  \end{array}$

  \caption{Example of results estimated by the SSCDNet trained with data augmentation of change mask. The left images $I_1$, $I_2$ and $M'$ show inputs of the SSCDNet. The top and bottom images in the middle column show the prediction results by the SSCDNet trained without and with the data augmentation of change mask. The right images $L_1$, $L_2$ and $M$ show the ground-truth of the semantic labels and the change mask.  (The images $I_1$, $I_2$, $M$, $L_1$ and $L_2$ are parts of the Vistas dataset \cite{Neuhold_2017_ICCV}.)}
  \label{fig:sscdnet_test_mask_da}
  \end{center}
  \vspace{-2mm}
\end{figure}

\begin{figure*}[!t] 
  \begin{center}

    \input{fig_supp/tex/00000383.tex} 
     \vspace{-3mm}
    \hspace*{8mm}\raisebox{2mm}{\includegraphics[width=160mm,bb=0 0 824 1]{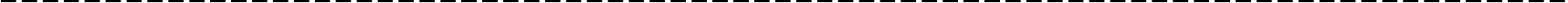}}        
     \vspace{0mm}
    \input{fig_supp/tex/00000389.tex} 
    \vspace{-3mm}
    \hspace*{40mm}\includegraphics[width=160mm,bb=0 0 1092 44]{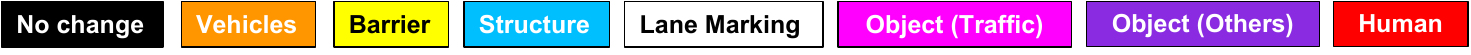}   
     \vspace{0mm}
    \caption{Examples of semantic scene change detection for the PSCD dataset. One failure case is shown in the lower part.}
    \label{fig:result_pscd}
  \end{center}
\end{figure*}

\begin{table*}[!t]
  \vspace{-4mm}
  \caption{IoU of the semantic change detection for the PSCD dataset.}
  \label{table:iou_pscd}
  \centering
  {\small
  \begin{tabular}{c|C{1.2cm}C{1.2cm}|C{1.2cm}C{1.2cm}|C{1.2cm}C{1.2cm}|C{2cm}}
    \hline
     &  \multicolumn{4}{|c|}{CSCDNet + SSCDNet}  & \multicolumn{2}{|c|}{GT mask + SSCDNet} & CSSCDNet \\
    \hline
    Training data (CD / SCD) & \multicolumn{2}{|c|}{PCD / Vistas}  & \multicolumn{2}{|c|}{PSCD (mask) / Vistas} & \multicolumn{2}{|c|}{ - / Vistas} & PSCD (full) \\
    \hline
    DA for training & - &\checkmark & - &\checkmark & - &\checkmark & n/a \\
    \hline
    mIoU & 0.192 & \textbf{0.196} & 0.215 & \textbf{0.223} & \textbf{0.303} & 0.283 & 0.322 \\
    \hline
  \end{tabular}
  }
  \vspace{-4mm}
\end{table*}

\subsubsection{Accuracy of the SSCDNet for synthetic data}
\label{subsubsec:accuracy_sscd}
Figure \ref{fig:sscdnet_test} shows an example of the results estimated using the SSCDNet.
The SSCDNet can accurately estimate semantic changes on each input image even if there are overlapping areas of change between input images.
Table \ref{table:iou_sscdnet} shows the mIoU of the SSCDNet for the synthetic validation data from the Mapillary Vistas dataset.
There are four combinations of training and test datasets with or without the data augmentation of the aforementioned change mask.
In the case of test data without data augmentation, namely, the input change mask is quite accurate, the SSCDNet trained using the dataset without the data augmentation performs better than one trained with the augmentation.
However, in the case of test data with the augmentation, namely, the input change mask has some errors, the SSCDNet trained using the augmentation outperforms the other.
Figure \ref{fig:sscdnet_test_mask_da} shows an example of results estimated by the SSCDNet trained with the data augmentation of the change mask.
The estimation results obtained using the SSCDNet trained without the data augmentation of the change mask have errors due to the errors from the input change mask.
However, the SSCDNet trained with the augmentation can accurately predict the semantic change labels while being more robust to the effects of errors of the input change mask.

\subsubsection{Semantic change detection for the PSCD dataset}
\label{sec:exp_pscd}
Figure \ref{fig:result_pscd} shows examples of the semantic change detection results for the entire process of our proposed method.
Table \ref{table:iou_pscd} shows the mIoU of each method for the PSCD dataset.
In Fig.\ref{fig:result_pscd}, the CSCDNet trained with the PCD dataset can accurately detect scene changes, although some detection errors are owing to occlusion of vegetation and small advertisement boards on the buildings because of the lack of training data. 
(The CSCDNet trained with the PSCD dataset can also detect them.)
Certainly, if the semantic change detection dataset, of which the creation is labor-intensive, is available, the strategy of the end-to-end learning for semantic change detection can be applied, and the performance is almost the best (CSSCDNet).
However, even if the dedicated dataset is unavailable, the SSCDNet can estimate semantic scene changes for each input image successfully depending on the change detection accuracy.
The lower part of Fig.\ref{fig:result_pscd} shows failure cases due to a lack of training data.

Better performance was achieved when the CSCDNet was trained using the PSCD dataset rather than being trained on the PCD dataset, which indicates that only the change detection dataset of the same domain as the target data should be used if it is available.
Furthermore, the SSCDNet using the ground-truth change mask performs close to the CSSCDNet, which is trained using the full set of the semantic change detection dataset.
Hence, the SSCDNet will exhibit high performance when accurate change mask information is available by other methods \cite{Sakurada2013,Taneja2011,Taneja2013} and sensors.

\section{CONCLUSIONS}
\label{sec:conclusion}
We proposed a novel semantic change detection scheme with only weak supervision.
The proposed method is composed of the two CNNs, the CSCDNet and the SSCDNet.
The CSCDNet can deal with the difference of camera viewpoints and achieves state of the art change detection performance for the PCD dataset.
The SSCDNet can be trained with dataset synthesized from semantic image segmentation datasets to avoid creating a new dataset for semantic change detection.
To evaluate the effectiveness of the proposed method, we created the first publicly available street-level image dataset for semantic scene change detection, named as the PSCD dataset.
Experimental results with this dataset verified the effectiveness of the proposed scheme in the semantic change detection task.

\section*{ACKNOWLEDGMENT}

This work is partially supported by KAKENHI 18K18071 and the New Energy and Industrial Technology Development Organization (NEDO).

\bibliographystyle{IEEEtran}
\bibliography{egbib}

\end{document}